\documentclass{article}
\usepackage{spconf,amsmath,graphicx}

\usepackage{hyperref}
\usepackage{amsmath,amssymb,amsfonts}
\usepackage{booktabs}
\usepackage{balance}  
\usepackage{url}  

\usepackage[linesnumbered,ruled,vlined]{algorithm2e}  

\SetKwInput{KwData}{Input}
\SetKwInput{KwResult}{Output}

\title{Measuring Class-Imbalance Sensitivity of Deterministic Performance Evaluation Metrics}
\name{Azim Ahmadzadeh and 
        Rafal A. Angryk}
\address{Department of Computer Science, Georgia State University, USA}

\begin{document}
%
\maketitle
\begin{abstract}
    The class-imbalance issue is intrinsic to many real-world machine learning tasks, particularly to the rare-event classification problems. Although the impact and treatment of imbalanced data is widely known, the magnitude of a metric's sensitivity to class imbalance has attracted little attention. As a result, often the sensitive metrics are dismissed while their sensitivity may only be marginal. In this paper, we introduce an intuitive evaluation framework that quantifies metrics’ sensitivity to the class imbalance. Moreover, we reveal an interesting fact that there is a logarithmic behavior in metrics’ sensitivity meaning that the higher imbalance ratios are associated with the lower sensitivity of metrics. Our framework builds an intuitive understanding of the class-imbalance impact on metrics. We believe this can help avoid many common mistakes, specially the less-emphasized and incorrect assumption that all metrics’ quantities are comparable under different class-imbalance ratios.
    
\end{abstract}
\section{Introduction}\label{sec:intro}
    
    One of the intrinsic challenges in forecast problems is the fact that the events of interest are scarce. This scarcity results in a large skew in the data, a phenomenon known as the \textit{class-imbalance issue}.
    This phenomenon seeks special treatments in order to avoid distorted analyses; magnified unfounded patterns and overly optimistic performances. Most of the existing treatments take place at two different levels. One is carried out at the data level, by changing the class distribution in order to create a balanced dataset \cite{drummond2003c4,chawla2002smote}. The other takes place at the model level, by adjusting the learning cost functions
    or the learning strategy \cite{sun2007cost}. Regardless of the approach taken, a test dataset should most closely mimic the actual characteristics of the data, including the scarcity of its classes. This takes us to the challenge of evaluating models' performance given skew data. Understanding the impact of skewness of data on evaluation is crucial as a metric may `misjudge' a model's performance due to the imbalance of the data \cite{hossin2015review}. This raises two important questions: (1) which metrics are completely insensitive to the class-imbalance issue, and (2) how can we determine, or even better, quantify the sensitivity of a metric. While there seems to be a deep understanding of the former in the community (e.g., \cite{krawczyk2016learning} among many) 
    , the latter seems to have attracted little attention.
    
    \begin{figure}[t]
        \centerline{\includegraphics[width=0.9\linewidth]{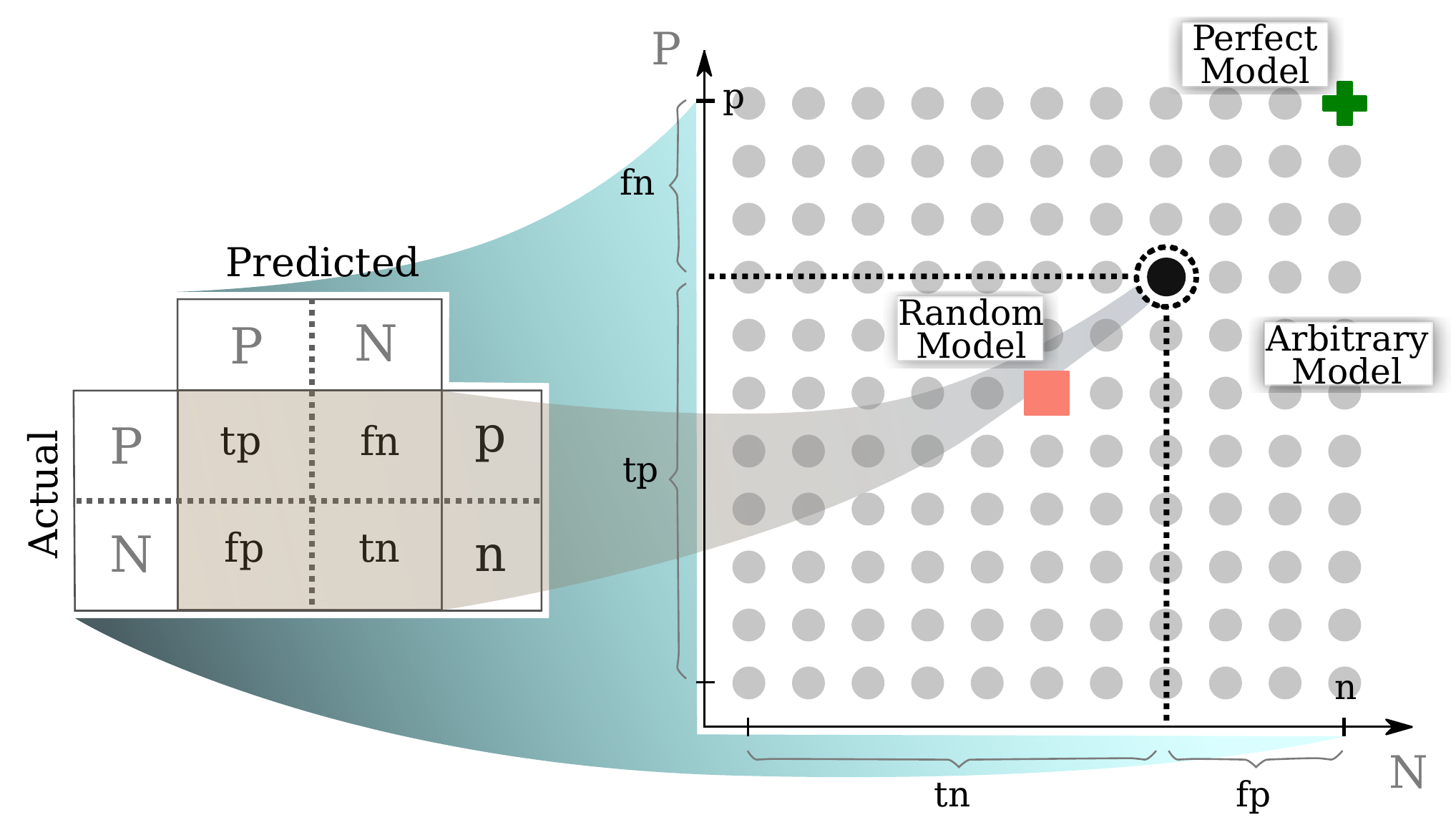}}
        \caption{Construction of the \textit{base contingency space} inspired by the ROC space. Each point in this space represents a family of \textit{relatively identical} confusion matrices.}
        \label{fig:cover_image}
        \vspace{-0.4cm}
    \end{figure}
    
    This broad understanding of the impact is not complete without the ability to measure the magnitude of the impact. This missing piece of knowledge causes limitations to data-driven studies as researchers may immediately dismiss a performance evaluation metric due to its known class-imbalance sensitivity, despite the fact that it might capture the strengths and weaknesses of their models very well. Since the magnitude of this sensitivity is often unknown to the researchers, they will never know whether the impact of this sensitivity is negligible or significant. From a different perspective, knowing the exact impact is also beneficial for critiquing research articles. A fair and thorough assessment of a research study based on the written report is a challenging task. This is primarily due to the existence of several critical factors which are not explicitly presented to the reviewer. Knowing the sensitivity of the employed evaluation metric(s) can help in this direction and allows a more accurate and less subjective assessment of the study.


\section{Background}\label{sec:background}
    The Receiver Operating Characteristic (ROC) curve \cite{peterson1954theory,swets1973relative} has been the predominant evaluation metric in almost all quantitative domains. It is simply a mapping of two variables, $tpr$ (true-positive rate, $\frac{tp}{p}$) and $fpr$ (false-positive rate, $\frac{fp}{n}$). The 2$D$ space formed by $fpr$ and $tpr$ as the $x$ and $y$ axes, respectively, is called the ROC space. A very similar space (with a slight modification as discussed in Sec.~\ref{sec:surfaces_and_sensitivity}) is illustrated in Fig.~\ref{fig:cover_image}. A trained, supervised model $m_i$ with a specific returned confusion matrix is mapped to the point defined by its true-positive and false-positive rates, i.e., the point $(fpr_i, tpr_i)$ in this space. There are three special points in this space which are reserved for \textit{trivial classifiers}: $(0,0)$ represents the \textit{all-negative} model\textemdash the model which classifies all samples as negative; $(1,1)$ stands for the \textit{all-positive} model\textemdash the one that classifies all samples as positive; and $(0,1)$ points to the \textit{perfect} model\textemdash the one that correctly classifies all positive and negative instances. The point $(1,0)$ would also represent the perfect model if the class labels were reversed. In addition to these special points, any model whose corresponding point on the ROC space lies on the main diagonal (where $tpr\!=\!fpr$) has no discriminative skill, and therefore is called a \textit{no-skill} model. These geometrical settings give ROC space the unique interpretability and intuitiveness that has made it so popular.
    
    
    One of the direct benefits of defining the ROC space is to help find the optimal value of a model's parameter $\theta$, e.g., a biomarker in diagnosis of a disease, or a hyper-parameter of a classifier. The simplest criterion for approaching this optimality problem is the distance from $(0,1)$. The model closest to this point is considered to be the most effective and therefore, the corresponding $\theta$ is considered optimal. A similar but less popular criterion is to maximize the distance from the anti-diagonal line. This makes it possible for different models to be compared after they are stripped of the success they achieve by chance. When the \textit{vertical} distance from the anti-diagonal line is employed, this criterion becomes a metric that is previously known as the \textit{Youden's $J$ index} \cite{youden1950index}. For probabilistic models ROC curve was utilized by setting different thresholds on the returned probabilities in order to obtain as many points as needed to form a curve in the ROC space. The Area Under the ROC Curve (AUC) of different models can then be used as a proxy for comparing probabilistic models' overall performance \cite{beck1986use, deLong1988comparing, swets1988measuring}. This idea was expanded to Volume Under the ROC Surface (VUS) for multi-class problems \cite{ferri2003volume}. A simpler expansion of this idea was by averaging AUC of all possible pairwise combination of the classes, known as \textit{micro averaging} \cite{hand2001simple}. To overcome one of the shortcomings of ROC, which is its disregard for any a priori knowledge about the classification costs when comparing different models, ROC Convex Hull (ROCCH) was introduced, based on the observation that the optimal model lies on the edge of the convex hull of the ROC curve \cite{provost2001robust}. This idea was also expanded to non-binary problems \cite{srinivasan1999note}. As pointed out in \cite{hand2009measuring}, AUC of different models are not comparable because its value depends on the cutoffs that themselves are related to the class imbalance of data. The \textit{cost space} \cite{drummond2000explicitly} and \textit{precision-recall space} \cite{davis2006relationship} are the proposed alternatives to mitigate the evaluation bias imposed by the class-imbalance of data.
    
    Taking advantage of the intuitiveness of the ROC space, we devise our new metric that quantifies the sensitivity of deterministic metrics to the class imbalance.

\section{Metric Surfaces and Imbalance Sensitivity}\label{sec:surfaces_and_sensitivity}
    
    \noindent\textbf{Relatively Identical Performances.}
        As we reviewed in Sec.~\ref{sec:background}, any given confusion matrix can be mapped to a single point in the ROC space. This mapping, however, is non-injective. That is, for an arbitrary point in the ROC space, represented by a pair $(fpr, tpr)$, there are infinitely many corresponding binary confusion matrices. The confusion matrices in each of these families are practically identical except that they reflect the performance given different sample sizes and/or different class-imbalance ratios of data. For performance evaluation, however, only their relative values are meaningful. This relative form is $\langle \frac{tp}{p}, \frac{fn}{p}, \frac{tn}{n}, \frac{fp}{n} \rangle$ which, for a given imbalance ratio $r$, can be simplified to a function of only two variables, $tpr$ and $tnr$, and can be reformulated as $\!\langle tpr, 1\!-\!tpr, r\cdot tnr,r(1\!-\!tnr) \rangle\!$. Since we want to introduce our sensitivity metric as a function of the imbalance ratio (hence independent of the sample size), we define this family of confusion matrices as follows: given a class-imbalance ratio $r$, we say two or more confusion matrices are \textit{relatively identical} if and only if they are identical in their relative forms.

            
        
    \noindent\textbf{ROC as a Bounded Semi-metric Space.} 
        ROC can be seen as a bounded semimetric space \cite{wilson1931semi}. Recall that a \textit{metric space} is a set $X$ that is endowed with a distance metric $d$. And $d\!:\!X^2\! \to \! \mathbb{R}$ is a \textit{distance metric} if, and only if, for all $a,b,c\!\in\! X$ the following conditions hold: (1) $d(a,b)\!\geq\! 0$, (2) $d(a,b)\!=\!d(b,a)$, and (3) $d(a,b)\!\leq\! d(a,c)\! +\! d(c,b)$. A distance metric that does not necessarily hold the third condition (triangle inequality) is called a \textit{semimetric}. In ROC space, $d$ can be the Euclidean distance between the points representing confusion matrices. In literature, using this distance metric, models' performance is often measured with their distance from the perfect model, i.e., $(0,1)$. In this study however, we do not explore the effectiveness of different distance metrics. Instead, we show that ROC's geometrical setting allows mapping of an arbitrary metric (that is derived from a confusion matrix) to a unique surface in an expansion of this space.
    
        Before we proceed, we need to make a small tweak in the ROC space to introduce symmetry. We alter ROC's $x$ axis (representing $fpr$) with $tnr$ (i.e., $1\!-\!fpr$). In other words, we flip the ROC space horizontally. Although the flipped ROC remains topologically identical to ROC, this transformation plays a critical role. Because the origin of this space (i.e., $(0,0)$) and the perfect model (i.e., $(1,1)$) do not share an axis any more, and the perfect model is placed farthest from the origin, the flipped ROC can be expanded to higher dimensions addressing multi-class evaluation. By adding more dimensions, the perfect model remains at the farthest place relative to the origin, which preserves the desirable spatial characteristic of the space. This is not possible in the ROC space. We call this flipped space the \textit{base contingency space} (illustrated in Fig.~\ref{fig:cover_image}).
        
        
        \begin{figure}[t]
            \centerline{\includegraphics[width=1.0\linewidth]{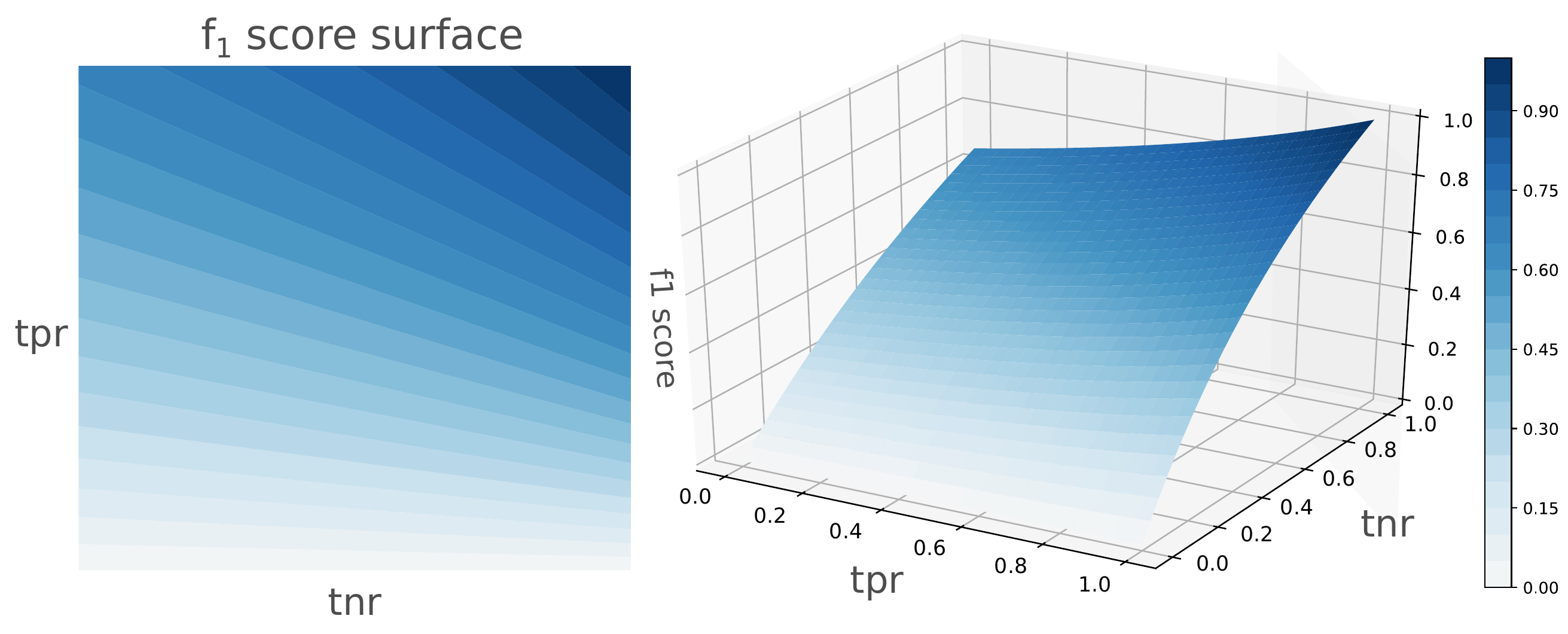}}
            \caption{The metric surface of $f_{1}$ score visualized in the contingency space. On the left, a contour plot is used to illustrate the surface whereas on the right, the surface's 3D structure is visualized for comparison.}
            \label{fig:surface_example}
            \vspace{-0.4cm}
        \end{figure}
        
    \noindent\textbf{Contingency Space and Metric Surfaces.}
        Deterministic performance evaluation metrics (e.g., accuracy, $f_{\beta}$ score) are functions, $\mu : [0,1]^2 \to \mathbb{R} $, that convert relatively identical confusion matrices (points in $[0,1]^2$) to single values (in $\mathbb{R}$), and therefore provide easy-to-interpret quantities. Considering the base contingency space as the domain of such functions, and by re-scaling the range of these functions to $[0,1]$, we can form a 3D space $[0,1]^3$ in which each metric $\mu$ can be represented uniquely by the surface it forms. We refer to this 3D space as the \textit{Contingency Space}, and call the surface corresponding to each metric a \textit{metric surface}.
        
        In Fig.~\ref{fig:surface_example}, an example of such a metric surface is illustrated. The one on the left is the contour plot of the surface representing precision in the contingency space, and the one on the right is its actual 3D visualization. The color tones in the contour plot represent the third dimension, i.e., the models' performance measured by the given metric. All plots in this paper are generated with the Python plotting library, \textit{matplotlib} v3.1.2 \cite{Hunter:2007}.

        \begin{figure}[t]
            \centerline{\includegraphics[width=0.8\linewidth]{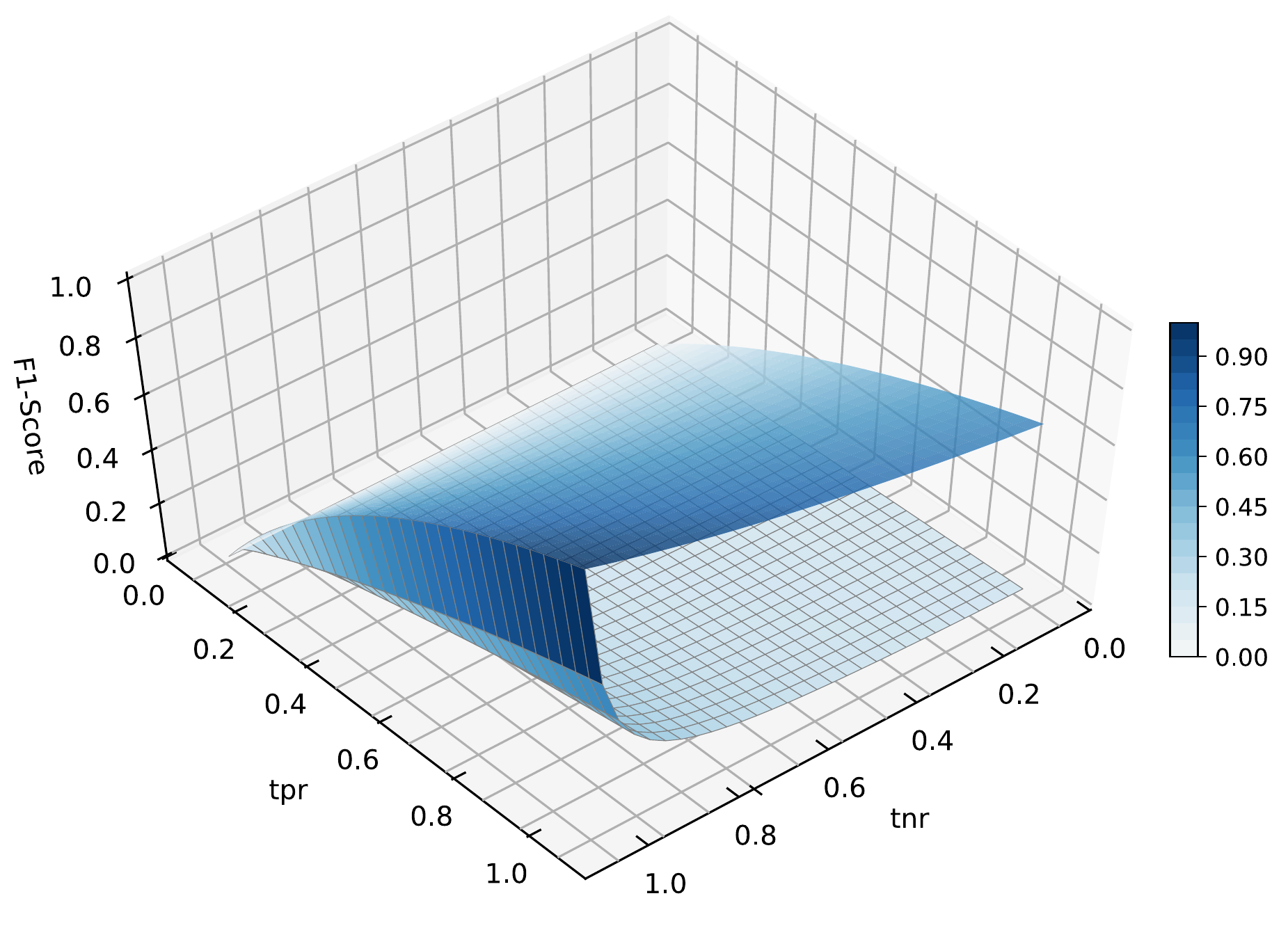}}
            \caption{Two metric surfaces of $f_{1}$ score are visualized in the contingency space. The top surface is formed assuming the imbalance ratio of 1:1, and the bottom surface is formed assuming the imbalance ratio of 1:49.}
            \label{fig:volume}
            \vspace{-0.4cm}
        \end{figure}
        
    \noindent\textbf{Imbalance Sensitivity.}
        Performance metrics are often sensitive to the class-imbalance ratio.
        The bivariate distribution of models' performance, imposed by an imbalance-sensitive metric, changes as the imbalance ratio in the data changes. The introduced contingency space captures this sensitivity very well. Fig.~\ref{fig:volume} provides an example of such a change for the metric $f_1$ score. In this figure, two surfaces are shown. The upper one corresponds to $f_1$ score with a 1:1 balance, whereas the lower one is generated with an imbalance ratio of 1:49. As we increase the imbalance ratio the surface warps quite drastically. As evident in this example, the $f_1$ score under this imbalance condition assigns high values only to models with very high $tnr$ values. For not-too-high values of $tnr$, on the other hand, the values of $tpr$ seem almost irrelevant to the $f_1$ score's evaluation. For a balance case, however, both $tpr$ and $tnr$ values have roughly equal weights in determining models' performance in terms of $f_1$ score. The difference between the two surfaces can only be attributed to the class-imbalance ratio. Inspired by this observation, we can quantify the degree of which a metric is sensitive to the class imbalance of data.
        
        \begin{figure*}[t]
            \centerline{\includegraphics[width=1.0\linewidth]{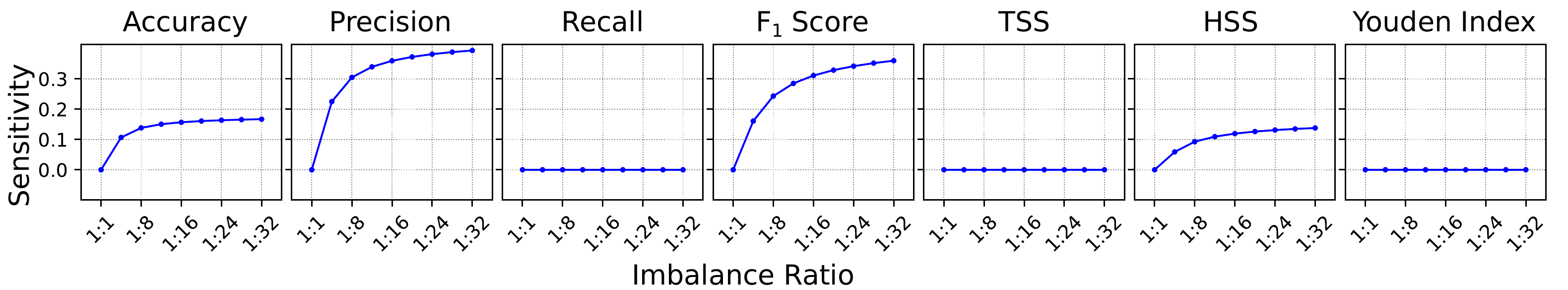}}
            \caption{Class-imbalance sensitivity of different performance evaluation metrics measured using the introduced contingency space and the surfaces metrics form in this space.}
            \label{fig:sensitivity}
            \vspace{-0.4cm}
        \end{figure*}
        
        Let $\{D_i\}_{i=1}^{n}$ denote a set of $n$ subsets of the dataset $D$, each with an arbitrary imbalance ratio. Suppose the model $m$ performs identically across all $n$ subsets. That is, it classifies a fixed percentage of the positive and negative instances correctly for all subsets, yielding identical confusion matrices for all $n$ cases. We say a metric $\mu$ is \textit{imbalance agnostic} if it returns a constant value as the model's performance for all $n$ subsets. We shall formulate this condition by $\mu(m(D_i))_{i=1}^{n} = \mathcal{C}$, where $\mathcal{C} \in \mathbb{R}$ is a constant, and both $m$ and $\mu$ are considered to be functions; $m$ takes in a dataset and returns a confusion matrix, and $\mu$ takes in a confusion matrix and returns a numeric evaluation that reflects $m$'s performance.

        While it might seem that an arbitrary metric does return a constant value as $m$'s performance on all $D_i$s, majority of the metrics in fact do not pass this condition. Accuracy is the most popular example of this kind. It is a function of the class-imbalance ratio. An example of an imbalance-agnostic metric, on the other hand, is recall ($\frac{tp}{p}$). Since it only measures the relative frequency of true positives, it is not a function of the imbalance ratio and therefore, it is imbalance agnostic.

        Given the above definition, it should now be clear that metric surfaces of imbalance-agnostic metrics have a unique characteristic which is that they do not warp at all as the imbalance ratio increases. With this intuition, we can now quantify a metric's sensitivity to the class imbalance by the magnitude its surface warps as a function of the imbalance ratio.
        
        Let $C^{1}(\mu)$ and $C^{r}(\mu)$ denote the surfaces formed by the metric $\mu$, with the class-imbalance ratios 1:1 and 1:$r$ ($r\!\neq\!1)$, respectively. The \textit{class-imbalance sensitivity} of $\mu$, denoted by $s_{\mu}^{r}$, is defined as the volume confined between $C^{1}(\mu)$ and $C^{r}(\mu)$. Since this volume depends on the imbalance ratio $r$, class-imbalance sensitivity becomes conveniently a function of $r$. This is very useful as in many rare-event analyses, some potentially useful metrics are excluded only because they are not completely insensitive to the imbalance ratio. However, the imbalance ratio of the data might be small and therefore it may impact a sensitive metric only marginally. Knowing the degree of sensitivity as a function of the imbalance ratio makes a strong argument for avoiding such exclusions.
        
        Note that the upper bound for the class-imbalance sensitivity is the volume of the (bounded) contingency space itself, which is equal to $1$. Therefore, the class-imbalance sensitivity is desirably limited to the range $[0,1)$. Also, it is worth mentioning that we can re-define the imbalance agnostic concept using this measure; a metric is imbalance agnostic if, and only if, $\forall\!r\!\in\!\mathbb{R}^{+}, \: s_{\mu}^{r} = 0$.
        
        The plots of imbalance sensitivity for a few deterministic performance metrics are shown in Fig.~\ref{fig:sensitivity}. The metrics accuracy ($\frac{tp + tn}{p + n}$), precision ($\frac{tp}{p'}$), recall ($\frac{tp}{p}$), and $f_1$ score ($2\cdot\frac{pre\cdot rec}{pre + rec}$ ) need no explanation as they are very popular across many domains. In these formulas, $p$ and $n$ are the total number of positive and negative instances, and $p'$ represents the total number of instances classified as positive (whether correctly or incorrectly). $tss$ measures the difference between the probability of detection and the probability of false alarm, i.e., $tss\!=\!\frac{tp}{p} - \frac{fp}{n}$ \cite{hanssen1965relationship}. The updated Heidke skill score ($hss$) \cite{balch2008updated} quantifies models' performance by comparing their performance with the random-guess model, and is formulated as $hss=\frac{2 ((tp \cdot tn) - (fn \cdot fp))}{p (fn + tn) + n (tp + fp)}$. Both $tss$ and $hss$ are popular metrics for rare-event classification tasks such as weather forecasting. Youden's $J$ index, is defined as $\frac{tp\cdot tn - fn\cdot fp}{(tp + fn)\cdot(fp + tn)}$ \cite{youden1950index}. Recall's sensitivity is indeed zero, in line with what we discussed above. Similarly, $tss$ and Youden's $J$ index are imbalance agnostic. Precision appears to be the most sensitive one in our list, followed by $f_1$ score, accuracy, and $hss$, respectively.
        
        Our proposed sensitivity measure reveals an interesting fact that sensitivity of metrics has a logarithmic growth. We have observed this pattern in several other metrics as well (e.g., Gilbert’s success ratio \cite{gilbert1884finley}, Doolittle index \cite{doolittle1888association}). We know that using imbalance sensitive metrics, two models' performance are comparable if, and only if, the imbalance ratios of the data used for evaluation of the models are identical. What we further learn from this logarithmic behavior is that the sensitivity impact becomes less and less severe as the imbalance ratios increases, and therefore, the sensitive metrics become comparable again.
        
        
    \noindent\textbf{Computational Complexity Analysis.}
        The time complexity of the class-imbalance sensitivity measure is constant. This can be broken down into two parts; representation of surfaces and computation of the volume confined between them. A metric surface can be represented by a square matrix with constant number of rows/columns, hence $O(1)$. To compute the imbalance sensitivity, we approximate the confined volume by the Riemann Sum i.e., $s_{\mu}^{r}\!=\!\sum_{i=1}^{t} \sum_{j=1}^{t} | c_{ij}^{1}(\mu) \!-\! c_{ij}^{r}(\mu) |$. Again, since the size of these surfaces are irrelevant to the sensitivity of their corresponding metrics, this part is also $O(1)$.

\section{Conclusion and Future Work}\label{sec:conclusion}
    In this study, we introduced the contingency space using which any metric that is derived from the confusion matrix can be mapped to a unique surface. If a metric's surface does not transform as the class-imbalance ratio changes, the metric is not sensitive to the imbalance of data. Otherwise, if the surface warps, the degree of the impact is proportional to the degree of this change.
    As our future work, we believe tracking the learning progress of a model on the surface of each metric provides richer context for the performance evaluation analysis. Moreover, the contingency space is expandable to higher-dimensional spaces which could be directly used as a metric that measures models' performance without having to aggregate multiple binary evaluations. We hope that the proposed measure assists researchers in finding the appropriate metrics they need for evaluation of their models.
    
\section*{Acknowledgment}
    This project has been supported in part by funding from CISE, MPS and GEO Directorates under NSF award \#1931555, and by funding from the LWS Program, under NASA award \#NNX15AF39G.
    
\vfill\pagebreak

\bibliographystyle{IEEEbib}
\bibliography{strings,refs}

\begin{thebibliography}{10}

\bibitem{drummond2003c4}
Chris Drummond, Robert~C Holte, et~al.,
\newblock ``C4. 5, class imbalance, and cost sensitivity: why under-sampling
  beats over-sampling,''
\newblock in {\em Workshop on learning from imbalanced datasets II}. Citeseer,
  2003, vol.~11, pp. 1--8.

\bibitem{chawla2002smote}
Nitesh~V Chawla, Kevin~W Bowyer, Lawrence~O Hall, and W~Philip Kegelmeyer,
\newblock ``Smote: synthetic minority over-sampling technique,''
\newblock {\em Journal of artificial intelligence research}, vol. 16, pp.
  321--357, 2002.

\bibitem{sun2007cost}
Yanmin Sun, Mohamed~S Kamel, Andrew~KC Wong, and Yang Wang,
\newblock ``Cost-sensitive boosting for classification of imbalanced data,''
\newblock {\em Pattern Recognition}, vol. 40, no. 12, pp. 3358--3378, 2007.

\bibitem{hossin2015review}
Mohammad Hossin and MN~Sulaiman,
\newblock ``A review on evaluation metrics for data classification
  evaluations,''
\newblock {\em International Journal of Data Mining \& Knowledge Management
  Process}, vol. 5, no. 2, pp. 1, 2015.

\bibitem{krawczyk2016learning}
Bartosz Krawczyk,
\newblock ``Learning from imbalanced data: open challenges and future
  directions,''
\newblock {\em Progress in Artificial Intelligence}, vol. 5, no. 4, pp.
  221--232, 2016.

\bibitem{peterson1954theory}
WWTG Peterson, T~Birdsall, and We~Fox,
\newblock ``The theory of signal detectability,''
\newblock {\em Transactions of the IRE professional group on information
  theory}, vol. 4, no. 4, pp. 171--212, 1954.

\bibitem{swets1973relative}
John~A Swets,
\newblock ``The relative operating characteristic in psychology: a technique
  for isolating effects of response bias finds wide use in the study of
  perception and cognition,''
\newblock {\em Science}, vol. 182, no. 4116, pp. 990--1000, 1973.

\bibitem{youden1950index}
William~J Youden,
\newblock ``Index for rating diagnostic tests,''
\newblock {\em Cancer}, vol. 3, no. 1, pp. 32--35, 1950.

\bibitem{beck1986use}
J.~R. Beck and E.~Shultz,
\newblock ``The use of relative operating characteristic (roc) curves in test
  performance evaluation.,''
\newblock {\em Archives of pathology \& laboratory medicine}, vol. 110 1, pp.
  13--20, 1986.

\bibitem{deLong1988comparing}
E.~DeLong, D.~DeLong, and D.~Clarke-Pearson,
\newblock ``Comparing the areas under two or more correlated receiver operating
  characteristic curves: a nonparametric approach.,''
\newblock {\em Biometrics}, vol. 44 3, pp. 837--45, 1988.

\bibitem{swets1988measuring}
John~A Swets,
\newblock ``Measuring the accuracy of diagnostic systems,''
\newblock {\em Science}, vol. 240, no. 4857, pp. 1285--1293, 1988.

\bibitem{ferri2003volume}
C{\'e}sar Ferri, Jos{\'e} Hern{\'a}ndez-Orallo, and Miguel~Angel Salido,
\newblock ``Volume under the roc surface for multi-class problems,''
\newblock in {\em European conference on machine learning}. Springer, 2003, pp.
  108--120.

\bibitem{hand2001simple}
David~J. Hand and Robert~J. Till,
\newblock ``A simple generalisation of the area under the {ROC} curve for
  multiple class classification problems,''
\newblock {\em Mach. Learn.}, vol. 45, no. 2, pp. 171--186, 2001.

\bibitem{provost2001robust}
Foster Provost and Tom Fawcett,
\newblock ``Robust classification for imprecise environments,''
\newblock {\em Machine learning}, vol. 42, no. 3, pp. 203--231, 2001.

\bibitem{srinivasan1999note}
Ashwin Srinivasan,
\newblock ``Note on the location of optimal classifiers in n-dimensional roc
  space,''
\newblock 1999.

\bibitem{hand2009measuring}
David~J. Hand,
\newblock ``Measuring classifier performance: a coherent alternative to the
  area under the {ROC} curve,''
\newblock {\em Mach. Learn.}, vol. 77, no. 1, pp. 103--123, 2009.

\bibitem{drummond2000explicitly}
Chris Drummond and Robert~C Holte,
\newblock ``Explicitly representing expected cost: An alternative to roc
  representation,''
\newblock in {\em Proceedings of the sixth ACM SIGKDD international conference
  on Knowledge discovery and data mining}, 2000, pp. 198--207.

\bibitem{davis2006relationship}
Jesse Davis and Mark Goadrich,
\newblock ``The relationship between precision-recall and {ROC} curves,''
\newblock in {\em Machine Learning, Proceedings of the Twenty-Third
  International Conference {(ICML} 2006), Pittsburgh, Pennsylvania, USA, June
  25-29, 2006}, William~W. Cohen and Andrew~W. Moore, Eds. 2006, vol. 148 of
  {\em {ACM} International Conference Proceeding Series}, pp. 233--240, {ACM}.

\bibitem{wilson1931semi}
Wallace~Alvin Wilson,
\newblock ``On semi-metric spaces,''
\newblock {\em American Journal of Mathematics}, vol. 53, no. 2, pp. 361--373,
  1931.

\bibitem{Hunter:2007}
J.~D. Hunter,
\newblock ``Matplotlib: A 2d graphics environment,''
\newblock {\em Computing in Science \& Engineering}, vol. 9, no. 3, pp. 90--95,
  2007.

\bibitem{hanssen1965relationship}
AW~Hanssen and WJA Kuipers,
\newblock {\em On the Relationship Between the Frequency of Rain and Various
  Meteorological Parameters. (With Reference to the Problem of Objective
  Forecasting.)},
\newblock Koninklijk Nederlands Meteorologisch Instituut, 1965.

\bibitem{balch2008updated}
Christopher~C. {Balch},
\newblock ``{Updated verification of the Space Weather Prediction Center's
  solar energetic particle prediction model},''
\newblock {\em Space Weather}, vol. 6, no. 1, pp. S01001, Jan. 2008.

\bibitem{gilbert1884finley}
Grove~Karl Gilbert,
\newblock ``Finley's tornado predictions.,''
\newblock {\em American Meteorological Journal. A Monthly Review of Meteorology
  and Allied Branches of Study (1884-1896)}, vol. 1, no. 5, pp. 166, 1884.

\bibitem{doolittle1888association}
MH~Doolittle,
\newblock ``Association ratios,''
\newblock {\em Bull. Philos. Soc. Washington}, vol. 7, pp. 122--127, 1888.

\end{thebibliography}

\end{document}